\documentclass[sigconf]{acmart}

\usepackage{hyperref}
\hypersetup{
    colorlinks=true,
    linkcolor=blue,
    filecolor=magenta,      
    urlcolor=cyan,
}

\AtBeginDocument{%
  \providecommand\BibTeX{{%
    \normalfont B\kern-0.5em{\scshape i\kern-0.25em b}\kern-0.8em\TeX}}}

\setcopyright{acmlicensed}\acmConference[KDD '24]{Proceedings of the 30th ACM SIGKDD Conference on Knowledge Discovery and Data Mining}{August 25--29, 2024}{Barcelona, Spain}
\acmBooktitle{Proceedings of the 30th ACM SIGKDD Conference on Knowledge Discovery and Data Mining (KDD '24), August 25--29, 2024, Barcelona, Spain}
\copyrightyear{2024}
\acmYear{2024}
\acmDOI{10.1145/3637528.3671628}
\renewcommand{\shortauthors}{Avishek Kumar and Tyson Silver}
\acmISBN{979-8-4007-0490-1/24/08}

\begin{document}

\title{Know, Grow, and Protect Net Worth: Using ML for Asset Protection by Preventing Overdraft Fees}

\author{Avishek Kumar}
\orcid{0000-0002-9911-3640}
\affiliation{%
  \institution{Intuit CreditKarma}
  \streetaddress{1100 Broadway}
  \city{Oakland}
  \state{California}
  \country{USA}
  \postcode{94607}
}
\email{avishek.kumar@creditkarma.com}

%\author{Ivelin Georgiev Angelov}
%\affiliation{%
%  \institution{Intuit}
%  \streetaddress{2600 Marine Way}
%  \city{Mountain View}
%  \state{California}
%\email{ivelin_angelov@intuit.com}

%\author{Kymm Kause}
%\affiliation{%
%  \institution{Intuit}
%  \streetaddress{2600 Marine Way}
%  \city{Mountain View}
%  \state{California}
%\email{kymm_kause@intuit.com}

\author{Tyson Silver}
\orcid{0009-0009-3835-5193}
\affiliation{%
  \institution{Lightcast}
  \streetaddress{232 N Almon St}
  \city{Moscow}
  \state{Idaho}
  \country{USA}}
\email{tyson.silver@lightcast.io}
\authornote{This work was done while the author was at Intuit.}

\thanks{Any opinions, findings, and conclusions or recommendations expressed in this material are those of the authors and do not necessarily reflect the views of our employers.}

\renewcommand{\shortauthors}{Avishek Kumar and Tyson Silver}

\begin{abstract}
  When a customer overdraws their bank account and their balance is negative they are assessed an overdraft fee. Americans pay approximately \$15 billion in unnecessary overdraft fees a year, often in \$35 increments; users of the Mint personal finance app pay approximately \$250 million in fees a year in particular. These overdraft fees are an excessive financial burden and lead to cascading overdraft fees trapping customers in financial hardship. To address this problem, we have created an ML-driven overdraft early warning system (ODEWS) that assesses a customer’s risk of overdrafting within the next week using their banking and transaction data in the Mint app. At-risk customers are sent an alert so they can take steps to avoid the fee, ultimately changing their behavior and financial habits. The system deployed resulted in a \$3 million savings in overdraft fees for Mint customers compared to a control group. Moreover, the methodology outlined here is part of a greater effort to provide ML-driven personalized financial advice to help our members know, grow, and protect their net worth, ultimately, achieving their financial goals. 

\end{abstract}





%%  \includegraphics[width=\textwidth]{sampleteaser}
%%  \caption{Seattle Mariners at Spring Training, 2010.}
%%  \Description{Enjoying the baseball game from the third-base
%%  seats. Ichiro Suzuki preparing to bat.}

%%\received{20 February 2007}
%%\received[revised]{12 March 2009}
%%\received[accepted]{5 June 2009}

\maketitle

\section{Introduction}

Overdraft fees largely fall on the most vulnerable customers, often living paycheck-to-paycheck, and are considered excessive and exploitative\cite{valenti_joe_overdraft_2022}.  Americans typically pay \$15 billion a year in avoidable overdraft fees. The same 9\% of customers pay 80\% of overdraft fees, often paying more than 10 overdraft fees in a year. Overdrafts can be considered a way to monetize accounts with small depository amounts\cite{cfpb_cfpb_2023}. Moreover, overdraft fees can discourage the most vulnerable customers from even participating in the banking system due to the high cost of banking by incurring these fees\cite{cfpb_cfpb_2013}.
\begin{figure*}[t]
	\includegraphics[width=5in]{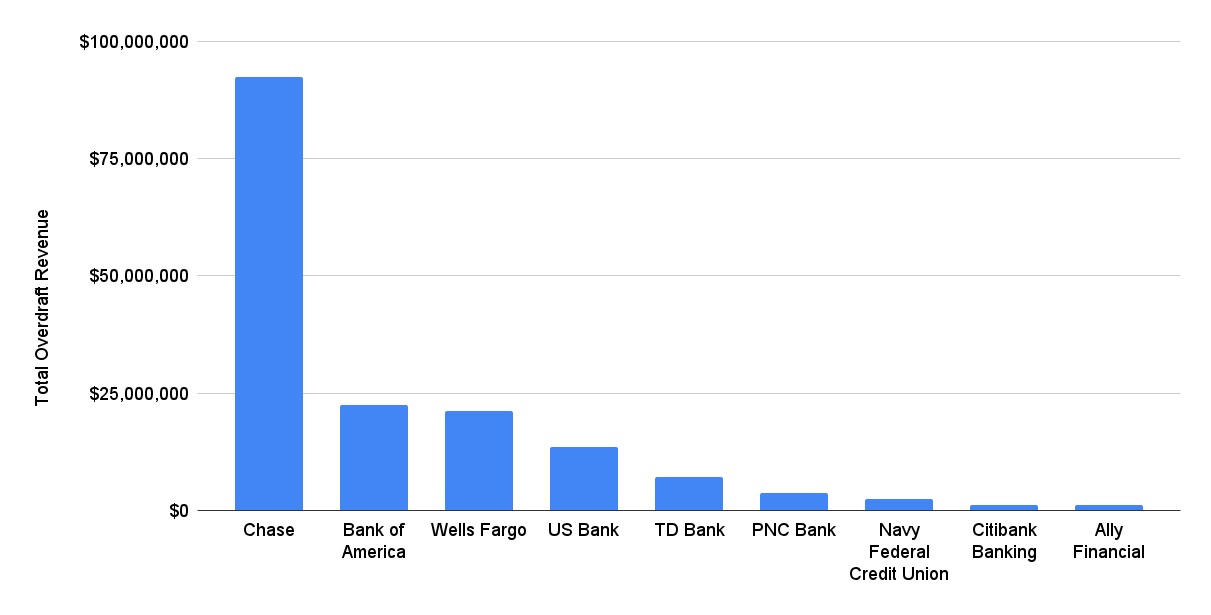}
    \caption{Overdraft Fees paid by Mint Customers per Bank from 09-2019--09-2020. On average Mint customers pay \$250 million in Overdraft Fees to banks often in \$35 increments. The share of overdraft fees collected by banks scales linearly with the number of customers. The goal of ODEWS is to provide a customer with an early warning to prevent paying unnecessary overdraft fees.  }
    \label{FIG:overdrafts_bank}
\end{figure*}
 
 An overdraft fee is charged by the bank for accepting a transaction when there are insufficient funds; a non-sufficient fund fee is charged for denying a transaction when there are insufficient funds. When an overdraft or non-sufficient fund fee is charged depends on a bank's terms of service. For this work we use the term overdraft fee for both types of fees as we are interested in preventing both.
 An account can be overdrawn in one of five ways due to insufficient funds: a check is denied (bounce a check), electronic debit, debit card transaction, ATM withdrawal, or in person withdrawal.    Originally, overdraft fees were a rare occurrence. The bank would charge a fee to customers who had not properly balanced their checkbook. With the popularization of debit transactions and electronic bill payments, overdraft fees have ballooned into a billion dollar revenue stream for banks. This is how a \$5 coffee transaction can become a \$40 transaction or a \$15 electronic bill payment can become a \$50 bill payment. Ultimately, this can lead to cascading overdraft fees, creating a financial burden, and trapping customers in a cycle of fees and low balances. 

 There has been a patchwork of temporary and uneven measures to combat overdraft fees. During the Covid-19 pandemic many banks granted waivers to customers who knew to request them while some banks waived overdraft fees for all customers temporarily.  There has also been legislation proposed in the United States, \textit{2021 Stop Overdraft Profiteering Act}\cite{booker_cory_stop_2022}, but not passed. AI tools mostly focus on financial markets--explainable AI for trading\cite{kumar_explainable_2022}, trading strategies\cite{jevtic_ai_2022}, risk management\cite{zhao_combining_2022}, and fraud detection\cite{varmedja_credit_2019, Tadapaneni2019-TADAII-2} rather than personal finance. 

 In  \cite{liu_analyzing_2018}, a Bayesian hierarchical model is proposed to find an optimal overdraft fee structure as a percentage of the overdrawn amount.    In  \cite{ramlal_personalized_2021}  an ML-framework for identifying how many overdrafts a customer should be allowed and what the cost should be is proposed.  There have also been retrospective economic studies on the impact of overdrafts and predatory lending\cite{melzer_competition_2015, alan_unshrouding_2018} and the positive benefits of providing SMS notifications for overdrafts and loan repayments\cite{caflisch_sending_2018}. In the behavioral sciences, there has been work on finding  effective nudge messaging to reduce overdrafts\cite{ben-david_using_2019}.
 
 % We would highlight that many banks already offer this type of service through opt-in overdraft protection. One such form of overdraft protection is where a customer can link their checking account to a credit card or line of credit. In this case, the credit limit serves as the limit on how many overdrafts a customer can cover and the APR of the credit card serves as the personalized interest rate charged to the customer for the overdraft. The credit limit and APR are established based on an already existing credit origination process that take into account personalized credit history. 

There are several challenges in preventing overdrafts fees that make it suitable for an ML-driven solution. Firstly, different banks have different policies regarding the cost of the overdraft fee, the number of overdraft fees that can be charged to a customer in a day, the grace period an account can be made current, different thresholds until an overdraft fee is assessed, and different fees for different types of overdrawing an account\cite{chang_ellenn_these_2020}. A ML-based model can learn these policies per bank and adapt to changes over time. Secondly, banks order transactions in order of largest amount transaction to lowest amount transaction during nightly processing (known as Transaction Processing Order) which can make it difficult to predict an overdraft. Although the Mint app receives balance and transaction data, the app does not receive the data in real time, oftentimes there can be a delay of hours to days depending on the financial institution. The transaction processing order and lag in receiving data rule out the possibility of simply waiting for a pending transaction to appear, calculating whether the transaction will place a customer in a negative balance, and then alerting the customer before the transaction is posted or using a time-series model. Instead, an ML-model can leverage financial data and predict if a customer is going to overdraft. Finally, we are seeking to quantify the risk of a customer overdrawing their account. There are a constellation of factors that lead to an overdraft besides a point-in-time balance and debits that can be leveraged in a systematic and structured way using an ML-model to provide an individualized measure of risk, ranking of customers based on risk, and even inform the type of intervention. 

\subsection{Our Contribution}
In light of temporary and uneven measures from banks, stalled legislative efforts, and other challenges in preventing overdraft fees, the authors have created an ML-driven Overdraft Early Warning System (ODEWS) on the Mint app.  The problem has been cast as a binary classification problem where the system predicts a customer’s likelihood of overdrawing their checking account in the next week by leveraging platform and transaction data. At-risk customers are then provided an early warning through email notification to prevent overdrafts and ultimately save our customer's money. This work builds on prior work of \cite{ben-david_using_2019} by focusing on the ML-aspects of assessing the risk of overdrafting and providing an effective intervention. In addition to theoretical models that propose percentage fee structures\cite{liu_analyzing_2018} and alternative overdraft pricing strategies \cite{ramlal_personalized_2021}, the ODEWS system focuses on preventing overdrafts and has been deployed, tested, and shown to successfully prevent overdrafts.   Moreover, the authors chose to focus on overdrafts due to their large impact on our customers, but this framework can be adopted for preventing other types of junk fees as well. An example of the output of the model is shown in Table \ref{table: output}, where customers are ranked by their risk of overdrafting. The risk score is the output of the model. The highest at-risk of overdrafting are then sent a notification to prevent the overdraft.

\begin{table}
\centering
\begin{tabular}{|p{.35cm}|p{1.3cm}|p{1.5cm}|p{1.5cm}|p{1.0cm}|p{.75cm}|}
\hline
ID & Total \newline Overdrafts & Overdrafted \newline this week & Days Since \newline Overdraft & Balance & Risk Score \\
\hline
01 & 95 & True & 2 &-\$700.00 & 99 \\
\hline
02 & 5 & False & 9  &\$11.30  & 81 \\
\hline
03 & 1 & False & 104 &\$700.00 & 74 \\
\hline
\end{tabular}
\caption{An example of the output of ODEWS. A list of customers ranked by risk of overdrafting is prioritized for intervention. This table provides an assessment of risk based on transaction history and balance data. }
\label{table: output}
\end{table}

\subsubsection{Scope of Overdraft Problem}Since the time of this work and current time of this writing several banks have reduced overdraft fees, adopted less punitive policies, and a few smaller banks have outright eliminated the fees\cite{arnold_chris_people_2022}. Despite this positive change, many of the largest banks are still charging overdraft fees and overdraft fees still remain an unnecessary drag on a customer’s cash flow. In 2019, overdraft revenue was estimated to be \$12 billion, now it is estimated to be a \$9 billion revenue generator\cite{cfpb_cfpb_2023}. In a recent survey by the Consumer Financial Protection Bureau\cite{cfpb_cfpb_2023}, low-income households are charged overdrafts fees (34\% of households with \$65,000 in income) at a much higher rate compared to higher-income households (10\% of households with \$175,000 in income); most households that have incurred an overdraft fee could have used a credit card if they had been given a warning. The authors, therefore, believe that the prevention of overdraft fees and junk fees in general are still an important part of asset protection. 

\subsection{Overdraft Problem for Mint Customers}
\begin{figure*}[t!]
	\includegraphics[width=5in]{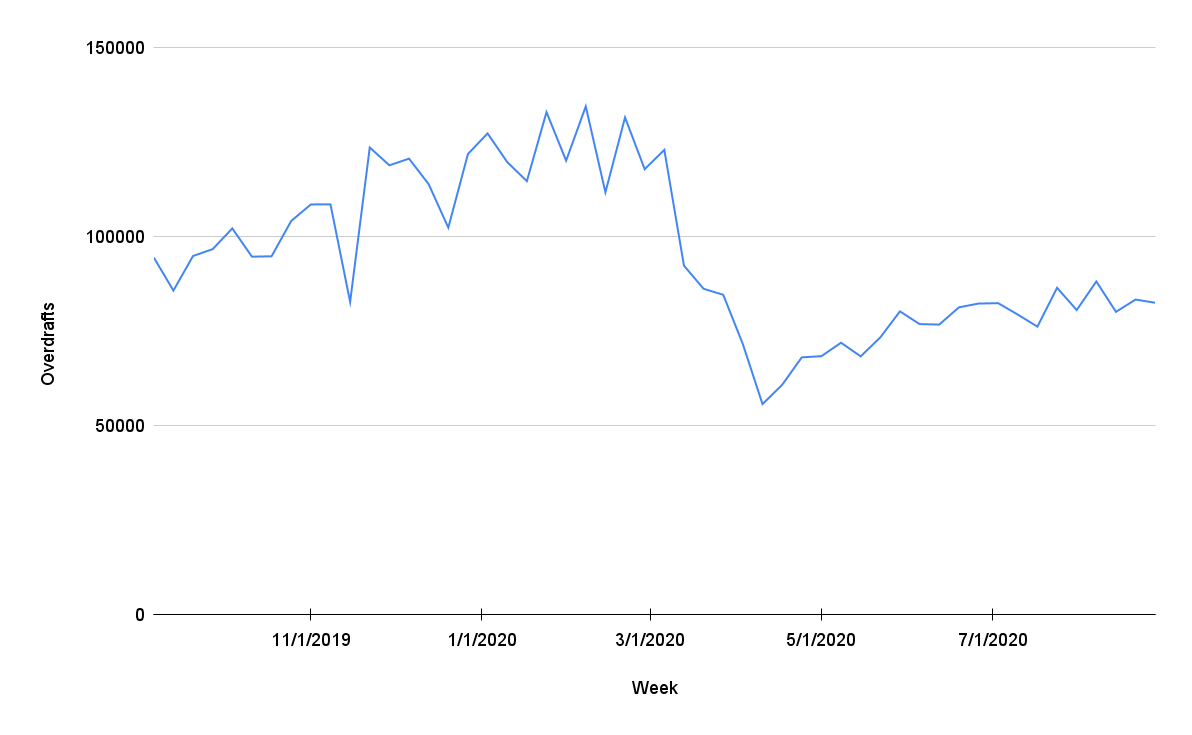}
    \caption{Overdrafts per Week from 09-2019--09-2020. The Covid-19 pandemic saw mass shutdowns which led to a precipitous decline in the number of overdrafts due to a precipitous decrease in the number of transactions as well as changes in bank overdraft policies. As the pandemic progressed the number of overdrafts gradually increased. The rapid change in overdrafts due to changing customer behavior as well as changing bank policies makes the problem of preventing overdrafts well suited to an adaptable machine-learning based solution.}
    \label{FIG:overdrafts_time}
\end{figure*}

The Mint app is a personal finance app with 20 million users in the United States and Canada that tracks customer’s transactions by linking their bank accounts, credit cards, and investment accounts; the app also provides budgeting tools, forecasting, and personal finance insights\footnote{As the time of this writing, the Mint app is currently being merged with the Intuit Credit Karma app where these features will be migrated.}. A decrease in unnecessary overdraft fees paid by Mint customers is a significant savings that protects our customer's cash flow--especially for those who are living paycheck-to-paycheck. Mint users pay on average \$250 million dollars in unnecessary overdraft fees every year. The average customer overdrafts 1.8 times before they are aware they overdrafted their account. Before this project, Mint customers would incur overdraft fees without warning. Figure \ref{FIG:overdrafts_bank} shows the breakdown of overdraft fees of Mint customers across nine banks over the period 9/2019-9/2020. Typically, the amount of fees collected scales with the number of customers with the largest three American banks, Chase, Bank of America, and Wells Fargo, collecting the largest share. 

ODEWS was deployed during the Covid-19 pandemic which saw a sudden and extreme shock on the American economy. Suddenly, many Americans were unemployed and businesses were shuttered, leading to a precipitous decline in economic activity.  As Figure \ref{FIG:overdrafts_time} shows, there was a sudden decrease in overdrafts at the beginning of the pandemic, approximately 03-2020, due to a sudden decrease in transaction volume as a result of shutdowns caused by the pandemic. Secondly, banks modified their overdraft policies to temporarily suspend overdrafts or adopt more lenient overdraft policies. In this case both the behavior of the customer as well as overdraft policies were changing lending itself to an ML-solution that is adaptable to a rapidly changing environment.

\section{Data Sources}
\begin{figure}
	\includegraphics[width=\linewidth]{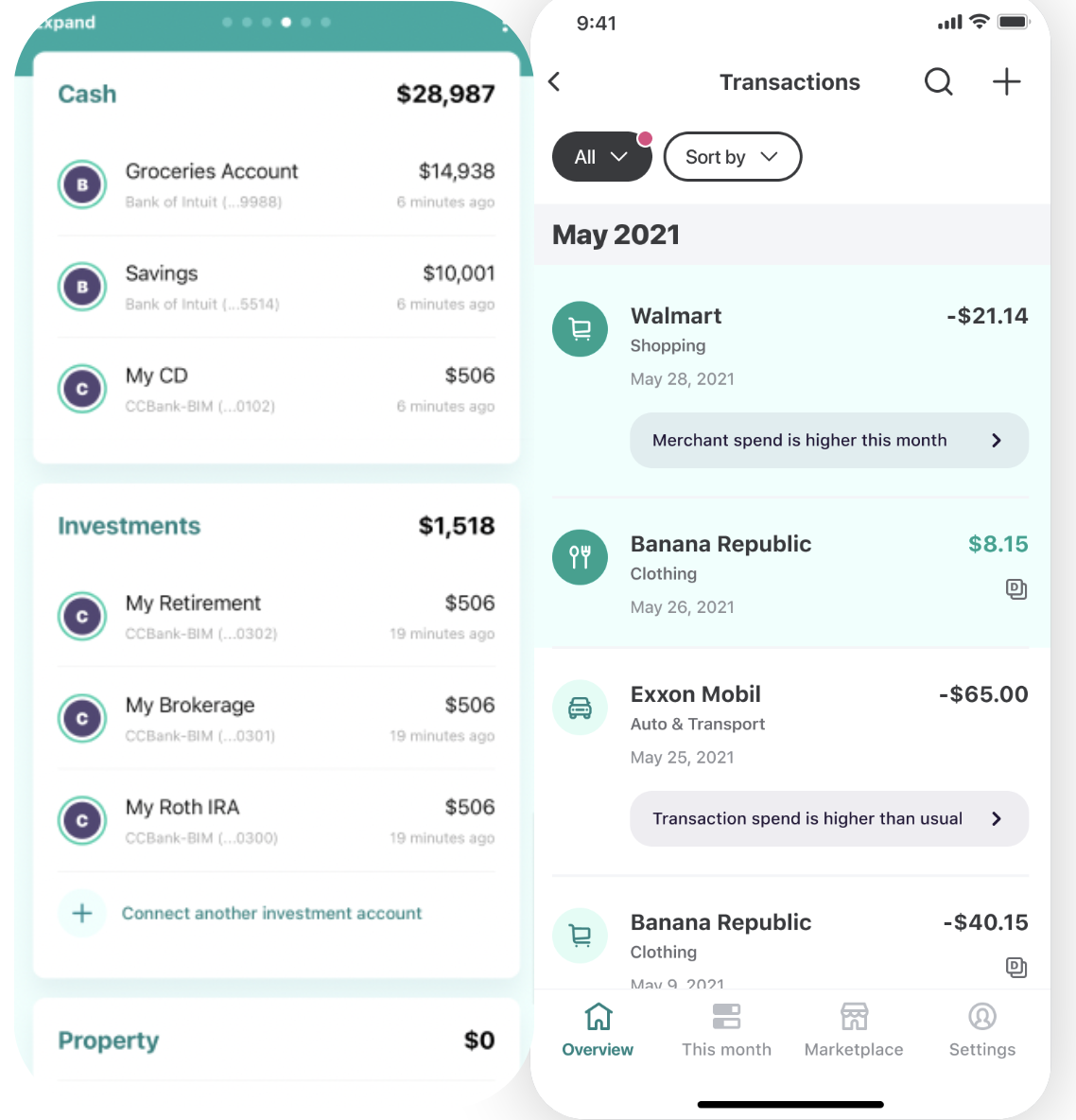}
    \caption{The Mint app is used for tracking a customer's transactions and managing their finances. Customers can link their savings, checking, and investments accounts and see their transaction history.}
    \label{mint_data}
\end{figure}

The data used for this work is derived from a customer’s banking, platform, and transaction data collected by the Mint app (Figure \ref{mint_data}). Customers can link their bank accounts, credit cards, investment accounts, and loans to the app to receive a panoramic view of their finances and receive financial insights. Data security policies can be found at \url{security.intuit.com}.  Mint has 3.6 million  monthly active users with transaction history for customers that can range from three months to over a decade depending on how long the customer has used the app. 

\subsubsection{Banking}
The Mint app allows customers to link their financial accounts. This includes checking and savings accounts, CDs, money market accounts, credit card accounts, and investment accounts as well as their respective balances.  

\subsubsection{Platform}
Platform data includes the number of logins a user has in the Mint app as well as clickstream regarding what pages they access in the app.  The number of logins is a measure of intent and proxy for how closely a customer monitors their finances\cite{chang_monitoring_2017,delgado_fuentealba_household_2021}. Our reasoning is that if a customer has logged into Mint recently they have likely assessed their finances and therefore at lower risk of overdrafting.

\subsubsection{Transaction History}
As part of linking financial accounts to the app, Mint will fetch the transaction history of customers. This includes the transaction timestamp, amount, transaction description, and transaction category.

\subsubsection{Data Limitations}
The main limitation of the data is due to the velocity the Mint app receives the latest banking and transaction data. The time from when a transaction is made and transits to the Mint app can be anywhere from hours to days. Several banks require customers to log-in to Mint, provide a two-factor authentication, and then the app can fetch the latest transactions. Moreover, several banks do not provide data over the weekend while transaction processing is occurring, which can lead to scenarios where a customer can spend over the weekend and not realize they have overdrafted until transaction processing is complete on Monday morning. Secondly, Mint bank balance data also has a lag that is often slower than the velocity that the app receives the latest transaction data. This can lead to scenarios where the app believes a customer has a positive balance but in actuality they have a negative balance and are actively overdrafting their account. 

\section{Methodology}
This section describes the overall methodology from extracting features for training, selecting and evaluating machine learning models. The unit of prediction is a customer and checking account on a given day. The majority of customers have a single checking account. There are a set of features derived from the customer as well as a set of features derived from the transaction history of the checking account.  

The label of whether a customer has overdrafted is derived by searching through transaction history. Each bank has their own unique transaction descriptions to indicate whether a fee has been assessed.  Searching through transaction history through manual curation and keyword search, we were able to identify the transaction descriptions that indicate an account was overdrawn and match the description to a bank based on the origin of the account. On the (simulated) day of prediction, we look-forward one week to search through customer transaction history and find any overdraft fee transactions. If any overdraft fees are found then this is a positive label (overdrafted within the next week) otherwise the customer/account is given a negative label (did not overdraft in the next week). 

The features of the model can be divided in several feature groups: overdraft features, transaction features, bank features, and login features. Overdraft features are related to a customer's history of overdrafting and are the most important features of the model. Generally, if a customer has ovedrafted before they are at a high risk of overdrafting again, especially if they have already ovedrafted in the same week. The overdraft features are the number of ovedrafts in the last six months in the account, number of overdrafts across all accounts in the last six months, days since the last ovedraft, amount of the last ovedraft, and whether there has been an overdraft in the last week. The bank features provide a panoramic view of how many accounts a customer has indicating a level of financial complexity. The bank level features are the number of checking accounts, credit card accounts, savings accounts, CD accounts, and investment accounts on the day predictions are made. Transaction features are related to the amount of debit and credit transactions in a checking account as well as the amounts, the balance of a checking account, and changes in the bank account balance. The account activity is predictive of the risk of overdrafting. The account level features are the last known balance in the account before the day of prediction as well as transaction activity before the day of prediction. Due to fetching both transactions and balances at different times, customer balances can often be out of sync of the true balance. To address this problem a new balance is calculated by fetching the last known balance, transactions that were made after the last known balance timestamp, and calculating a new balance. The transaction history of an account is converted into features by first dividing transactions into credit and debits and creating the following features: the total number of credit and debit transactions, min, max, and average amounts over a four-week and one-week look back window from the day of prediction; the number of debit transactions under \$50 as well the number of credits over \$200 over the last two months from the day of prediction. Different look-back windows were tested for different features.  Looking back six months provides a comprehensive look of a customer’s overdraft history, including the number of overdrafts, days since the last overdraft, and number of days since last overdraft. If, say, a one-month look back window was used, the model would be unable to separate chronic overdrafters from occasional overdrafters which would increase our false positives. Additionally, there could be a customer that recently overdrafted but may have not overdrafted in the previous six months. The model would find this customer at medium-to-low risk looking back six-months and likely not be in the at-risk pool. The look-back windows for features reported here provided the best model performance.

\section{Modeling Approach and Results}
The cohort of interest that receives a prediction are Mint customers who have overdrafted at least once in the last six months and have a checking account from the following banks: Chase Bank, Bank of America, Wells Fargo, US Bank, TD Bank, Citibank, PNC Bank, Navy Federal Credit Union, and Ally Bank. (Note: At the time of this deployment Citibank and Ally Bank were charging customers overdraft fees. They have since eliminated overdraft fees at the time of this writing. Now, each bank will simply deny transactions at their discretion based on their respective overdraft policies\cite{noauthor_citibank_2022,noauthor_ally_2021}).  Many Mint customers likely have overdraft protection coverage where if an overdraft occurs, rather than being charged a fee, the bank will draw funds from the customer’s savings account, line of credit, or charge a credit card. In an ideal world, all customers would have overdraft protection to avoid overdrafts fees. This is not possible for several reasons. A customer may not have a savings account to link with a checking account; a customer may also not have a line of credit or credit card available to cover an overdraft. Moreover, it may not be financially prudent to charge an overdrawn amount to a credit card and pay the interest on the debt or associated fees.  We do not have insight into which customers have this type overdraft protection and which do not. To prevent sending overdrafts warnings to customers with overdraft protection we restrict our cohort to customers who have overdrafted in the last six months and assume they have no overdraft protection since they have received an overdraft fee. There is a trade off to this approach where any customer who overdrafts for the first time in the last six months will not be in the prediction cohort. Although we are missing a small portion of overdraft fees through this conservative approach, we are avoiding false positive notifications ensuring customers have faith in the system when they do receive a notification. If a customer were to enable overdraft protection while being in the cohort, they would decrease in risk as time went on and fall out of the high-risk list of customers. For new customers, we typically receive three months worth of transaction history from the bank when they sign-up and they would be subject to the same criteria. 

The goal of the model is to predict if a customer will overdraft within the next week; to that end, the model calculates predictions every Friday afternoon.  We currently do not make a point-of-time prediction after a transaction is made or a daily prediction due to technical limitations with how quickly the app receives transaction data.  Also due to limitations with fetching data from banks, the Mint app receives transaction at different cadences from banks leading to performance differences for each bank compared to historical offline testing. Moreover, even if our system could predict if a customer was going to overdraft that day or the next, the customer often cannot change their behavior quickly enough to prevent the overdraft (based on customer interviews). The system, therefore, focuses on predicting the risk of overdraft over a one week window of time to allow customers sufficient time to change their behavior, and the authors saw good performance of the model using this label. Focusing on a longer time window such as a two-week window may be too much time and a customer may not feel the urgency to take corrective action. After the model generates predictions, a list of at-risk customers is created and they are sent an email notifying them we believe they are about to overdraft. This then provides the customers the opportunity to adjust their spending, transfer money, or contact the bank for a waiver. Due to diverging overdraft policies of different banks, particularly during the height of the Covid-19 pandemic, a separate model was created for each individual bank. Initially, we explored having a single model to predict overdrafts for all banks but found the model was learning the overdraft policies of the largest banks, hurting the performance for customers belonging to smaller banks. Rather than trying different sampling strategies the authors decided to create a model per bank, particularly to catch acute changes to overdraft policies by individual banks. Empirically, this provided better performance. 

\subsubsection{Temporal Cross-Validation:} The model training scheme is designed to mimic the deployment process in order to have the most accurate performance results. The model is trained using temporal cross-validation to take into account temporal effects and serial correlations that affect customer behavior (features) and overdraft policies (labels)\cite{foster_big_2021}. As can be seen from Figure \ref{FIG:overdrafts_time} and previously stated, during the height of the Covid-19 pandemic there were extreme exogenous shocks in customer behavior and overdraft policies. The model is retrained every week to take into account any temporal effects such as shutdowns, re-openings, government stimulus payments\cite{flitter_emily_their_2020}, and changes in overdraft policies that would affect overdraft behavior. 
An example of temporal cross-validation is the following: if simulating making a prediction on 2020-06-14 the label is calculated between the dates 2020-06-07–2020-06-14 and features are calculated looking back six months from the date 2020-06-07. The test-set of the model is then calculated using a label from 2020-06-14–2020-06-21 and features looking back six months from 2020-06-14. In this way, train-test pairs are created every week over six months and used to train models. 

\begin{figure}[h]
	\includegraphics[width=\linewidth]{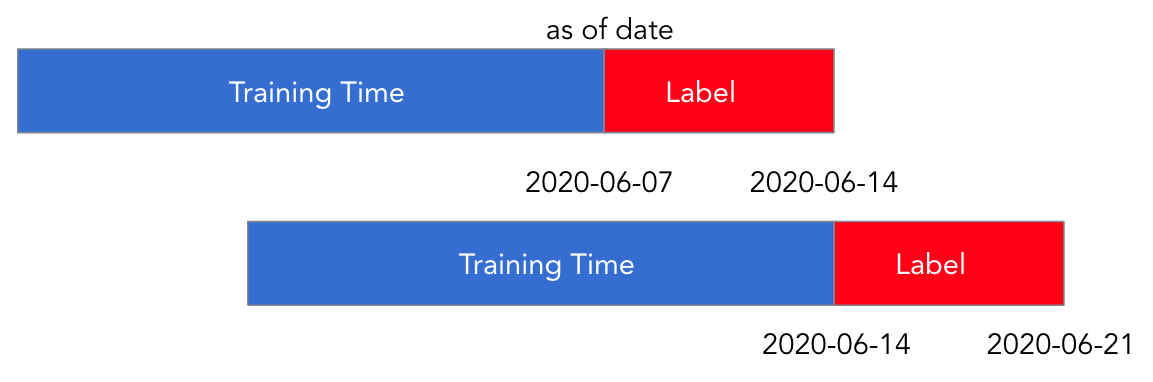}
    \caption{Temporal Cross Validation: when simulating making a prediction on 2020-06-14 the label is calculated between the dates 2020-06-07–2020-06-14 and features are calculated looking back six months from the date 2020-06-07. The test-set of the model is then calculated using a label from 2020-06-14–2020-06-21 and features looking back six months from 2020-06-14. In this way, train-test pairs are created every week over six months and used to train models taking into account temporal effects and serial correlations. }
    \label{temporal_cross_validation}
\end{figure}

\subsubsection{Model Selection:} The system has the following high-level requirements. 

\begin{enumerate}
    \item The system needs to be trusted by the user for it to be effective. In this context, the customer needs to understand intuitively why they received the message if they are a false positive so they have trust in the system.
    \item Our goal is to reach the largest group as possible to prevent overdrafts.
    \item The overdraft email notification is one of many different notifications customers receive (e.g., large expenses, payday, credit card offers, late fees, etc). A product requirement is to roughly keep the amount of notifications constant every week so one notification does not flood customers inbox over others, meaning there is a roughly fixed k of users that are notified.
\end{enumerate}

Different classification methods (Logistic Regression, Decision Trees, Random Forests Gradient Boosted Decision Trees, and Neural Nets), hyperparameters, training data time ranges, and feature sets were compared to each other using precision@k\%, recall@k\%\cite{kurzynski_evaluating_2018}, as well as model stability over time. A full model grid and hyperparameter set can be found in Table \ref{TABLE:modelgrid}.  Rather than optimizing the model for an aggregate metric such as AUC, the local precision-recall space is optimized to maximize the number of customers that will overdraft in the top k\% and provide an accurate measure of the performance of the deployed model. Precision@k\% and recall@k\% are metrics commonly found in information retrieval and search engine literature\cite{foster_big_2021} where it is important that the items flagged at the top of a list are accurate. It is appropriate for our requirements because we can only send roughly the same number of notifications each week, a requirement of the email system. Therefore, the system has to maximize the precision@k\% and recall@k\% of the top k\% to reach the largest audience. When using metrics such as precision@k\% and recall@k\%, NDCG (Normalized Discounted Cumulative Gain) is often calculated as well to quantitatively measure if the relevant customers or items are at the top of a predicted list. NDCG does not apply in this particular instance because we have no need to have a well-ordered list at k. All customers in our at-risk at k list regardless of where they are in the list will get an email notification.
 
The model informs which customers will receive an email intervention. An email is a low-cost intervention allowing considerable flexibility regarding who receives an email as well as how many emails can be sent every week. The only resource requirement is that the number of overdrafts messages is roughly constant each week due to there being a number of other email campaigns and the requirements of the system. This notification is one of many that customers receive (e.g., large expense, payday, credit offers, late fees etc). To guide how many notifications are sent we have to consider the costs of false positive and false negative cases. A false positive message can inure the customer to future messages and make them lose trust in the system. The risk of a false negative case is a customer does not receive a notification and overdraws their account. 

\begin{figure}[h]
	\includegraphics[width=\linewidth]{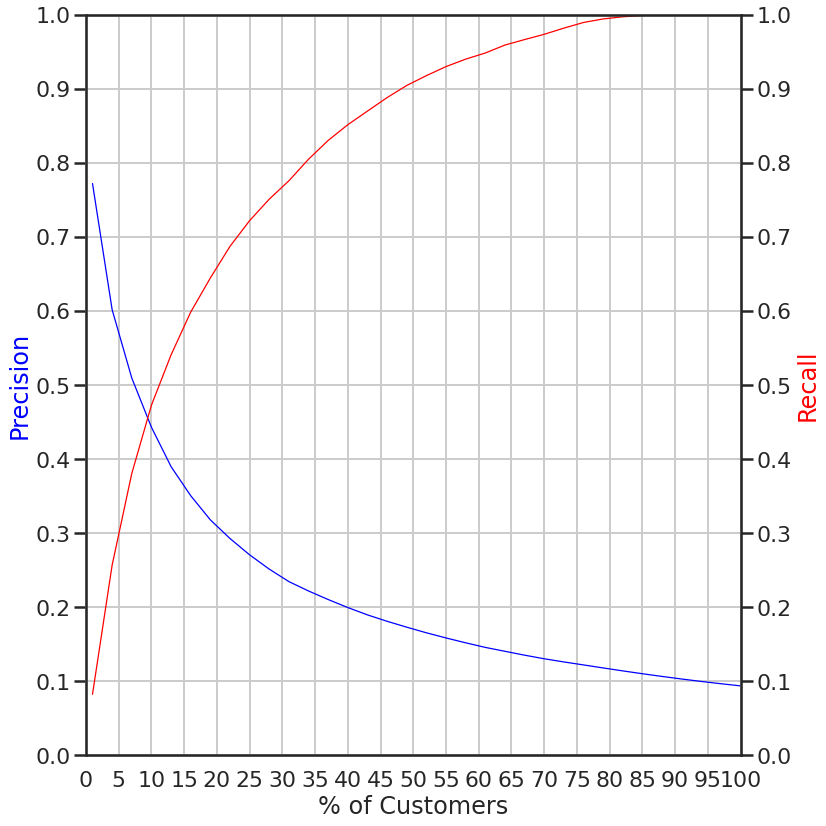}
    \caption{Precision and Recall @k\% for Chase Bank Model: The model informs who and how many people receive an email notification. There is considerable flexibility in how many notifications can be sent every week since there are no resource constraints. Models are selected trying to balance precision and recall. In this case the threshold of k\% is set at 10\% and precision@k\% is 0.42 and recall@k\% is 0.49, roughly balancing the precision@k\% and recall@k\%. The Chase Bank model has a 5x lift from the prior and 1.17x lift from business rules. }
    \label{p_n_r}
\end{figure}

Models were selected with the goal to maximize precision@k\% and recall@k\% at the highest k\%. To that end, models were selected where $recall@k\% \approx 0.4-0.5$ and $precision@k\% \approx 0.4-0.5$ to balance precision and recall. False positives and false negatives were implicitly balanced in this approach and assumed to be equally harmful. For the banks Wells Fargo, Chase Bank, US Bank, Navy Federal Credit Union, TD Bank and Citibank models that balanced precision@k\% and recall@k\% were found. In cases where precision@k\% and recall@k\% could not be easily balanced–Bank of America, PNC Bank, Ally Financial--precision@k\% was favored and set to $precision@k\% \approx 0.4-0.5$ to minimize false positives and maintain trust in the system. If the authors had maximized the recall@k\%, the large amount of false positive notifications would have led to a loss in trust in the system. The k\%, percentage of customers each week that receive a notification, precision@k\%, recall@k\% and model type can be found in Table \ref{TABLE: model_stats}. For completeness the AUC of the models used is reported in Table \ref{TABLE:ROCAUC} and the $precision@k$\% and $recall@k$\% are reported for Chase Bank across the different algorithms in Table \ref{TABLE:ChasePerformance}. Figure \ref{p_n_r}, shows a specific example for Chase Bank. In this case the threshold of k\% is set at 10\% and precision@k\% is 0.42 and recall@k\% is 0.45, roughly balancing the precision@k\% and recall@k\%. For 8 of the 9 banks, a GBDT (Gradient Boosted Decision Trees)\cite{friedman_stochastic_2002} model was the most performant model and the best performing model for Ally Financial was a Feed Forward Neural Net. The authors considered using a risk score cutoff or a top k cutoff for the model. In the end, we found both to be equivalent; the top k cutoff corresponded with a risk score cutoff and was stable over time.  Model hyperparameters and architecture can be found in Table \ref{TABLE:hyperparameters}. As is common in problems assessing risk, boosting largely performed better than deep learning methods\cite{rudin_secrets_2019}.

To understand if a sophisticated ML-system was worth deploying, each bank’s model was compared to the bank’s Bayesian prior of ovedrafting and a rules-based baseline. Overall, each model had a 4-15x lift compared to the prior. Since there is no current process for preventing overdrafts, a 2-deep decision tree was created for each bank to simulate the best business rule that can be found. Comparing each model to a 2-deep decision tree model, each model has a 1.17-2x lift compared to the 2-deep decision tree, showing the model was worth deploying.  Full details can be found in Table \ref{TABLE: model_stats}.

\begin{figure}[h]
	\includegraphics[width=\linewidth]{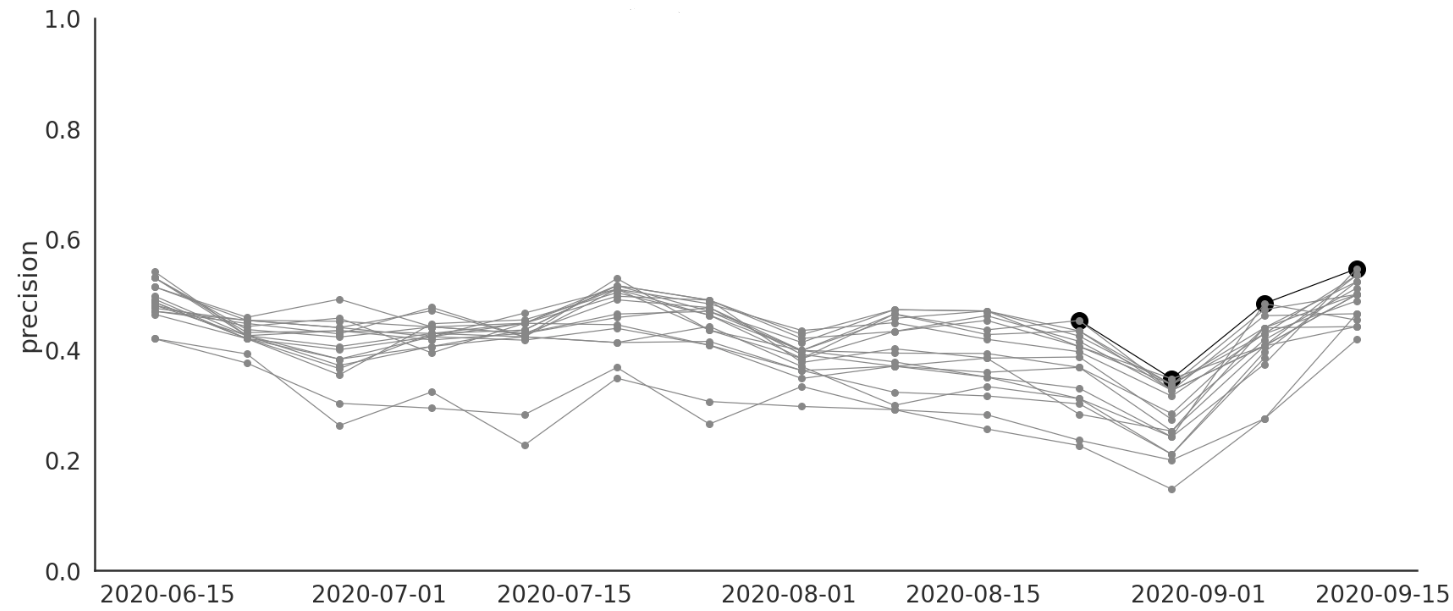}
    \caption{Balance Stability and Performance: Different models have different levels of variance and performance over time which can lead to large variances in performance week over week. Through temporal-cross validation we are able to measure the model performance over time after choosing how many customers, k\%, the model will flag. The final model selected is chosen for having the best performance over the last four weeks compared to the best possible performance of all models trained and being within 5\% of the precision of the best possible model within the last four weeks. }
    \label{stability}
\end{figure}

Another aspect of the model that was important to optimize is balancing performance and stability over time. As figure \ref{stability} shows different models had different levels of variance and performance over time which can lead to large variances in performance week over week. Through temporal-cross validation we are able to measure the model performance over time after choosing how many customers, k\%, the model will flag. The final model selected is chosen for having the best performance over the last four weeks compared to the best possible performance of all models trained and being within 5\% of the precision of the best possible model within the last four weeks. This method ensures both stability and performance.

\begin{figure}
	\includegraphics[width=\linewidth]{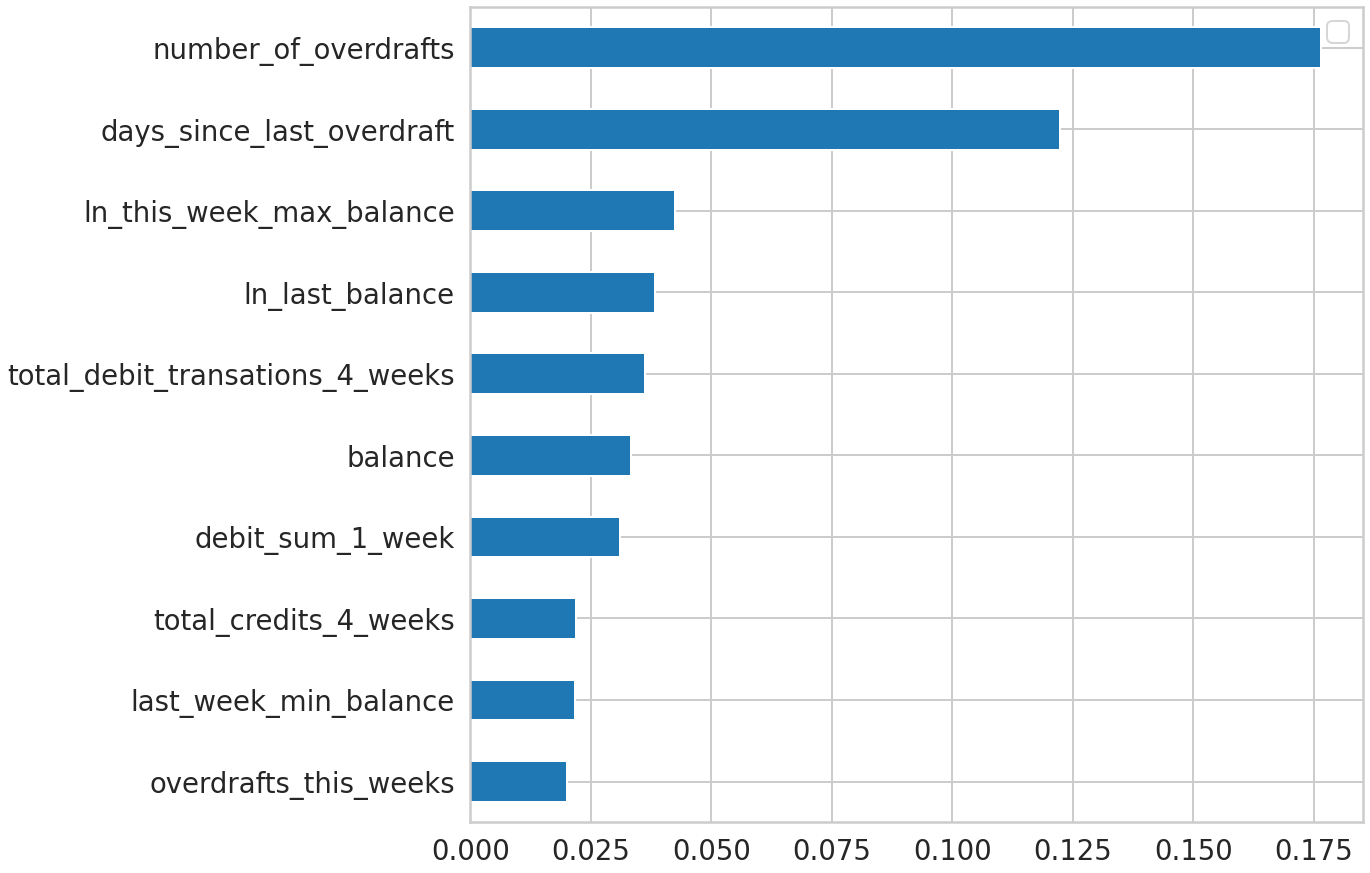}
    \caption{Feature Importances of Chase Overdraft Model. Feature importance is calculated using Gini importance. The top features of the Chase model (GBDT) are the number of times a customer has overdrafted, days since the last overdraft, whether they have overdrafted this week and the sum of debit transactions in the past week. These features are largely what was intuitively expected. If customers are regularly overdrafting, have low balances, or have already overdrafted in the week we expect the customer is at high risk of overdrafting again. }
    \label{feat_important}
\end{figure}

\begin{table*}[!ht]
    \centering
     \begin{tabular}{|l|l|l|l|l|l|l|l|}
    \hline
        Bank & Model Type & k\% & precision@k\% & recall@k\% & Prior (Baseline) & Lift over Prior & Lift over Business Rules \\ \hline
        Wells Fargo & GBDT & 10 & 0.43 & 0.45 & 0.1 & 4.3 & 1.23 \\ \hline
        Chase & GBDT & 10 & 0.42 & 0.49 & 0.09 & 4.67 & 1.17 \\ \hline
        Bank of America & GBDT & 6 & 0.46 & 0.33 & 0.09 & 5.11 & 1.44 \\ \hline
        Navy Federal Credit Union & GBDT & 7 & 0.45 & 0.5 & 0.05 & 9 & 2.05 \\ \hline
        US Bank & GBDT & 5 & 0.5 & 0.49 & 0.05 & 10 & 1.56 \\ \hline
        TD Bank & GBDT & 6 & 0.52 & 0.4 & 0.09 & 5.78 & 1.44 \\ \hline
        PNC Bank & GBDT & 2 & 0.45 & 0.26 & 0.03 & 15 & 1.88 \\ \hline
        Ally Financial  & FFNL & 2 & 0.48 & 0.17 & 0.06 & 8 & 1.66 \\ \hline
        Citibank Banking & GBDT & 6 & 0.46 & 0.4 & 0.07 & 6.57 & 1.44 \\ \hline
    \end{tabular}     \caption{Models were selected with the goal to maximize precision@k\% and recall@k\% at the highest k\%. To that end, models were selected where $recall@k\% \approx 0.4-0.5$ and $precision@k\% \approx 0.4-0.5$ to balance precision and recall.  For the banks Wells Fargo, Chase Bank, US Bank, Navy Federal Credit Union, TD Bank and Citibank models that balanced precision@k\% and recall@k\% were found. In cases where precision@k\% and recall@k\% could not be easily balanced–Bank of America, PNC Bank, Ally Financial--precision@k\% was favored and set to $precision@k\% \approx 0.4-0.5$ to minimize false positives. Each bank's model was compared to the bayesian prior baseline as well as a 2-deep decision tree to simulate a rules-based system. }
\label{TABLE: model_stats}
\end{table*}

\subsubsection{Feature Importances:}  Feature importances were calculated using Gini Importance\cite{Louppe:2013:UVI:2999611.2999660}. The top feature is the number of overdraft fees a customer has had in the past. This largely makes intuitive sense. Customers that are living paycheck-to-paycheck and have become accustomed to overdrafting are likely to continue overdrafting. Though, having a history of overdrafts is not solely predictive of overdrating an account in the next week.  There is a moderate correlation between this feature and other overdraft features such as average number of overdrafts each week. Secondly, the days since the last overdraft fee and if there has been an overdraft in the last week are also important. The average Mint customer overdrafts 1.8x before they realize they have overdrafted. The feature days since last overdraft was weakly correlated with all other features.  The model can often identify customers that have had a recent overdraft and alerts them before they have more. A large amount of debits indicates a highly active account where an overdraft is more likely. Debit features were moderately correlated with other debit and credit features. Customers typically spend after they have deposited into the accounts. Interestingly, the number of debit features were not correlated with the number of overdrafts. Our expectation was the accounts with the most active debits would be the most at-risk of overdrafts.  A low or already negative balance is also a high-risk indictator of an impending overdraft but not perfectly predictive. Certain banks will allow customers to hold a small negative balance, typically below \$50 dollars before an overdraft fee is assessed. Balance features were strongly correlated with other balance related features and moderately correlated with credit and debit features as would be expected because credit and debits are what affect the balance. On the low risk end, if a customer has low transaction activity in their account or a large number of days since their last overdraft they are at a much lower risk of overdrafting their account.

\section{Email Variants}
 Different variants of an email message were composed using nudge theory\cite{thaler_nudge_2021,cai_nudging_2020} and tested in order to find the most effective messaging, as measured by email open rates. Three variations of overdraft warnings were composed: “Overdrafts Happen. Here’s how to beat it” (Empowering Message), “Avoid paying overdraft fees of \$35 or more" (Loss Aversion Message), and “Overdraft Early Warning" (Baseline Message). As Figure \ref{FIG:email} shows, the subject line contains the nudge message. The body of the email contains the bank name of the account we believe the customer is at risk of overdrawing and the last time a customer received an overdraft fee. The bank name is provided so the customer knows what account we believe they will overdraw; for legal reasons that is the extent we can identify an account via email. Receiving a notification with the last overdraft fee date helps customers understand, intuitively, why they are receiving the message. There is also a list of actions we suggest to customers to avoid an overdraft.

\section{Model Test Design and Results}

The Overdraft Early Warning System was deployed and tested in an RCT (randomized control trial) to test the system’s ability to reduce overdraft fees. The null hypothesis is that the model cannot identify customers about to overdraft better than random and the email intervention has no effect on preventing overdrafts.  The test was run for 12 weeks from November 2020–January 2021. The model was retrained every Friday and subsequently used to generate a list of at-risk customers who were sent email notifications. For the test, customers were divided into four groups: control (at-risk but receive no notification) and a group for each of the three email variants, Baseline Message group, Loss Aversion Message group, and Empowering group. A customer is randomly assigned to one of the groups the first time they receive a notification and stay in that group for the duration of the test when receiving future notifications. Overall 200k notifications were sent and there were 60k participants in the test (48K in an email variant and 12k in control). As previously mentioned there is a lag in the velocity the Mint app receives the latest transaction data. In offline historical testing we measured a precision of 40\%-45\% for each bank's model (Table \ref{TABLE: model_stats}) where there is no data lag. We would therefore expect on average 45\% of customers in the test to overdraft each week during model deployment. In the control group, the percentage of customers that overdrafted each week was approximately 27\%. We attribute this performance difference to the lag in transaction data we receive at the time of prediction.  Notably, users in the treatment group who had never opened an email had similar overdraft rates as the control group. In these cases, emails from the Mint app may be summarily ignored or not reach the recipient (e.g., sent to 
 the spam filter). The overall email open rate across all three variants was 24\%. Of the users who opened an email, there was a 3.71\% (trending) reduction in customers overdrafting compared to the control group indicating the customer took some action or was mindful of overdrawing their account over the next week. Each email also contained a link to a user’s checking account on the Mint app. Of those users who clicked the link to see their balance, there is a 12.86\% reduction in customers overdrafting compared to the control group (statistically significant p < 0.05), indicating these users monitored their balances and likely took action to prevent overdrafts over the next week. 

An important goal of the system is to promote behavior change among customers that led to reducing overdrafts.  Users were tracked for three months from when they received their first email. Many certified financial planners suggest that it takes approximately three months for a new financial habit to develop.  Overall, there was an estimated savings of \$3 million dollars for customers in overdraft fees in the treatment group compared to control. There was also a \$5 difference in the average overdrafts fees per customer in the treatment group compared to the control group.

The purpose of creating and testing email variants is to find the most effective messaging. The results of the test support that customers who open the email are likely to have a reduction in overdrafts. A message that resonates and prompts the customer to action is therefore an important part of the system. The clear winner with a 30\% open rate is the Baseline Message, "Overdraft Early Warning",  compared to Empowering (21\%), "Overdrafts Happen Here’s how to beat it", and Loss Aversion (22\%), "Avoid paying overdraft fees of \$35 or more." The authors speculate that the simple message, “Overdraft Early Warning”, has a higher open rate because the message communicates that an overdraft is imminent and requires immediate action as opposed to the other variants that sound more like they are providing generic financial advice.

Another goal of the test was understanding the customer's ability to respond to an overdraft notification. The model predicted a customer's risk but cannot predict if a customer has the means to prevent an overdraft when they are notified. Typically, customers in the low-mid range of risk opened notifications more often compared to customers at high risk. As part of the test analysis customers were divided into three groups: high ability to respond, positive balances in accounts and few overdrafts; medium ability to respond, positive balances and many overdrafts; low ability to respond, negative balances in accounts and many overdraft. The high and medium ability to respond groups generally always had a smaller percentage of customers that overdrafted than the low ability to respond group as well as higher email open rates. We find that the low ability to respond group is not unaware they are overdrafting, rather they do not have the means to prevent overdrawing their accounts making this intervention ineffective for them. This raises the question of whether chronic overdrafters should receive email notifications even if we believe they cannot prevent the overdraft or will not interact with the notification. Given that the model exists in a social and ethical context, we believe that all customers should be warned of an overdraft regardless of their ability to prevent overdrafts. If certain customers receive email interventions and others do not, we risk creating or exacerbating a disparity\cite{foster_big_2021,saleiro_aequitas_2018} by not providing everyone at-risk the chance to prevent the overdraft.

Customer interviews during the test provided valuable insight into how customers react to the warning. One major concern was that false positives messages would lead to customers losing faith in the system and ignore future messages as a result.  In those interviews, we were able to learn that customers were not inured by false positives messages, email notifications that incorrectly identified customers that are about to overdraft, as long as they could intuitively understand why they received the notification based on their transaction and balance data.   Customers that received messages and could not avoid an overdraft were also still engaged and believed the notifications could be helpful when they had the ability to respond in the future.   Secondly, we found that no customer actually handled an overdraft the moment they received a notice and sometimes forgot until an overdraft fee was charged. We plan on adding an immediate push notification and later email reminder to be more effective in the future. A common request across interviews was for the Mint app to handle an impending overdraft automatically. In the future this could be done by automated or assisted movement of funds or requesting a waiver on the customer's behalf.

\section{Conclusions}
In this paper, the authors have presented a machine learning approach to develop a risk model for predicting which Mint customers will overdraft their checking account within the next week and provide an early warning email notification. Each bank’s model provides lift over the prior and outperforms a business rule model while also balancing stability and performance. Several message variations were tested using nudge theory to find the most effective messaging.
Overall, the system has saved customers \$3 million in overdraft fees during a test of the model. 

At a higher level, the methodology and approach can be used
to provide different types of financial advice. We have shown in this paper we can help customers avoid unnecessary overdraft fees. Although most ML-based work has focused on investing and financial markets, we strongly believe that ML-based tools can benefit personal finances. Specifically, we believe that we can provide ML-driven advice to help customers with other important aspects of personal finance such as increasing credit scores, building emergency savings, paying down debt, and allocating capital for investments. In the future, we hope to report on more ML-guided tools in personal finance.

\begin{acks}
 This work took a village to accomplish. I (AK) should like to acknowledge the following individuals: Ido Mintz, Itay Granik and Daniel Ben David for the original work around overdraft messaging; Victor Escoto for creating graphics for this paper; Joanne Locascio for developing the nudge messages; Kaitlin Inghilterra for coaching me on how to conduct a customer interview; Ivelin Angelov for helping to create the pipeline; Kymm Kause for creating dashboards for testing and analysis of the AB test; Ken Yocum and Kit Rodolfa for discussions and their careful review of this manuscript. 
\end{acks}

\bibliographystyle{ACM-Reference-Format}
\bibliography{odews}


\begin{thebibliography}{30}

%%% the Web address.

\ifx \showCODEN    \undefined \def \showCODEN     #1{\unskip}     \fi
\ifx \showDOI      \undefined \def \showDOI       #1{#1}\fi
\ifx \showISBNx    \undefined \def \showISBNx     #1{\unskip}     \fi
\ifx \showISBNxiii \undefined \def \showISBNxiii  #1{\unskip}     \fi
\ifx \showISSN     \undefined \def \showISSN      #1{\unskip}     \fi
\ifx \showLCCN     \undefined \def \showLCCN      #1{\unskip}     \fi
\ifx \shownote     \undefined \def \shownote      #1{#1}          \fi
\ifx \showarticletitle \undefined \def \showarticletitle #1{#1}   \fi
\ifx \showURL      \undefined \def \showURL       {\relax}        \fi
% invisible to TeX
\providecommand\bibfield[2]{#2}
\providecommand\bibinfo[2]{#2}
\providecommand\natexlab[1]{#1}
\providecommand\showeprint[2][]{arXiv:#2}

\bibitem[noa(2021)]%
        {noauthor_ally_2021}
 \bibinfo{year}{2021}\natexlab{}.
\newblock \bibinfo{title}{Ally {Bank} {Overdraft} {Policy}}.
\newblock
\newblock
\urldef\tempurl%
\url{https://www.ally.com/overdraft/}
\showURL{%
\tempurl}


\bibitem[noa(2022)]%
        {noauthor_citibank_2022}
 \bibinfo{year}{2022}\natexlab{}.
\newblock \bibinfo{title}{Citibank {Overdraft} {Policy}}.
\newblock
\newblock
\urldef\tempurl%
\url{https://online.citi.com/US/JRS/portal/template.do?ID=Citi-Overdraft-Fees-Change}
\showURL{%
\tempurl}


\bibitem[Alan et~al\mbox{.}(2018)]%
        {alan_unshrouding_2018}
\bibfield{author}{\bibinfo{person}{Sule Alan}, \bibinfo{person}{Mehmet
  Cemalcilar}, \bibinfo{person}{Dean Karlan}, {and} \bibinfo{person}{Jonathan
  Zinman}.} \bibinfo{year}{2018}\natexlab{}.
\newblock \showarticletitle{Unshrouding: {Evidence} from {Bank} {Overdrafts} in
  {Turkey}: {Unshrouding}: {Evidence} from {Bank} {Overdrafts} in {Turkey}}.
\newblock \bibinfo{journal}{\emph{The Journal of Finance}}
  \bibinfo{volume}{73}, \bibinfo{number}{2} (\bibinfo{date}{April}
  \bibinfo{year}{2018}), \bibinfo{pages}{481--522}.
\newblock
\showISSN{00221082}
\urldef\tempurl%
\url{https://doi.org/10.1111/jofi.12593}
\showDOI{\tempurl}


\bibitem[Arnold(2022)]%
        {arnold_chris_people_2022}
\bibfield{author}{\bibinfo{person}{Chris Arnold}.}
  \bibinfo{year}{2022}\natexlab{}.
\newblock \showarticletitle{People hate overdraft fees. {Banks} are ditching or
  reducing them.}
\newblock \bibinfo{journal}{\emph{NPR}} (\bibinfo{date}{Jan.}
  \bibinfo{year}{2022}).
\newblock
\urldef\tempurl%
\url{https://www.npr.org/2022/01/11/1071860136/people-hate-overdraft-fees-capital-one-is-ditching-them-and-other-banks-may-foll#:~:text=All%20U.S.%20banks%20together%20make,of%2010%20overdrafts%20a%20year.}
\showURL{%
\tempurl}


\bibitem[Ben-David et~al\mbox{.}(2019)]%
        {ben-david_using_2019}
\bibfield{author}{\bibinfo{person}{Daniel Ben-David}, \bibinfo{person}{Orly
  Sade}, {and} \bibinfo{person}{Ido Mintz}.} \bibinfo{year}{2019}\natexlab{}.
\newblock \showarticletitle{Using {AI} and {Behavioral} {Finance} to {Cope}
  with {Limited} {Attention} and {Reduce} {Overdraft} {Fees}}.
\newblock \bibinfo{journal}{\emph{SSRN Electronic Journal}}
  (\bibinfo{year}{2019}).
\newblock
\showISSN{1556-5068}
\urldef\tempurl%
\url{https://doi.org/10.2139/ssrn.3422198}
\showDOI{\tempurl}


\bibitem[Booker(2022)]%
        {booker_cory_stop_2022}
\bibfield{author}{\bibinfo{person}{Cory Booker}.}
  \bibinfo{year}{2022}\natexlab{}.
\newblock \bibinfo{title}{Stop {Overdraft} {Profiteering} {Act} of 2021}.
\newblock
\newblock
\urldef\tempurl%
\url{https://www.congress.gov/117/bills/s2677/BILLS-117s2677is.pdf}
\showURL{%
\tempurl}


\bibitem[Caflisch et~al\mbox{.}(2018)]%
        {caflisch_sending_2018}
\bibfield{author}{\bibinfo{person}{Andrea Caflisch},
  \bibinfo{person}{Michael~D. Grubb}, \bibinfo{person}{Darragh Kelly},
  \bibinfo{person}{Jeroen Nieboer}, {and} \bibinfo{person}{Matthew Osborne}.}
  \bibinfo{year}{2018}\natexlab{}.
\newblock \showarticletitle{Sending {Out} an {SMS}: {The} {Impact} of
  {Automatically} {Enrolling} {Consumers} {Into} {Overdraft} {Alerts}}.
\newblock \bibinfo{journal}{\emph{SSRN Electronic Journal}}
  (\bibinfo{year}{2018}).
\newblock
\showISSN{1556-5068}
\urldef\tempurl%
\url{https://doi.org/10.2139/ssrn.3538527}
\showDOI{\tempurl}


\bibitem[Cai(2020)]%
        {cai_nudging_2020}
\bibfield{author}{\bibinfo{person}{Cynthia~Weiyi Cai}.}
  \bibinfo{year}{2020}\natexlab{}.
\newblock \showarticletitle{Nudging the financial market? {A} review of the
  nudge theory}.
\newblock \bibinfo{journal}{\emph{Accounting \& Finance}} \bibinfo{volume}{60},
  \bibinfo{number}{4} (\bibinfo{date}{Dec.} \bibinfo{year}{2020}),
  \bibinfo{pages}{3341--3365}.
\newblock
\showISSN{0810-5391, 1467-629X}
\urldef\tempurl%
\url{https://doi.org/10.1111/acfi.12471}
\showDOI{\tempurl}


\bibitem[CFPB(2013)]%
        {cfpb_cfpb_2013}
\bibfield{author}{\bibinfo{person}{CFPB}.} \bibinfo{year}{2013}\natexlab{}.
\newblock \bibinfo{title}{{CFPB} {Study} of {Overdraft} {Programs}}.
\newblock
\newblock
\urldef\tempurl%
\url{https://files.consumerfinance.gov/f/201306_cfpb_whitepaper_overdraft-practices.pdf}
\showURL{%
\tempurl}


\bibitem[CFPB(2023)]%
        {cfpb_cfpb_2023}
\bibfield{author}{\bibinfo{person}{CFPB}.} \bibinfo{year}{2023}\natexlab{}.
\newblock \bibinfo{title}{{Overdraft} and {Nonsufficient} {Fund} {Fees}:
  {Insights} from the {Making} {Ends} {Meet} {Survey} and {Consumer} {Credit}
  {Panel}}.
\newblock
\newblock
\urldef\tempurl%
\url{https://files.consumerfinance.gov/f/documents/cfpb_overdraft-nsf-report_2023-12.pdf}
\showURL{%
\tempurl}


\bibitem[Chang et~al\mbox{.}(2017)]%
        {chang_monitoring_2017}
\bibfield{author}{\bibinfo{person}{Betty~P.I. Chang},
  \bibinfo{person}{Thomas~L. Webb}, \bibinfo{person}{Yael Benn}, {and}
  \bibinfo{person}{James~P. Reynolds}.} \bibinfo{year}{2017}\natexlab{}.
\newblock \showarticletitle{Monitoring personal finances: {Evidence} that goal
  progress and regulatory focus influence when people check their balance}.
\newblock \bibinfo{journal}{\emph{Journal of Economic Psychology}}
  \bibinfo{volume}{62} (\bibinfo{date}{Oct.} \bibinfo{year}{2017}),
  \bibinfo{pages}{33--49}.
\newblock
\showISSN{01674870}
\urldef\tempurl%
\url{https://doi.org/10.1016/j.joep.2017.05.003}
\showDOI{\tempurl}


\bibitem[Chang(2020)]%
        {chang_ellenn_these_2020}
\bibfield{author}{\bibinfo{person}{Ellenn Chang}.}
  \bibinfo{year}{2020}\natexlab{}.
\newblock \showarticletitle{These {Banks} {Are} {Waiving} {Overdraft} {Fees}
  {Because} of the {Coronavirus}}.
\newblock \bibinfo{journal}{\emph{U.S. News}} (\bibinfo{date}{March}
  \bibinfo{year}{2020}).
\newblock
\urldef\tempurl%
\url{https://money.usnews.com/banking/articles/these-banks-are-waiving-overdraft-fees-because-of-the-coronavirus}
\showURL{%
\tempurl}


\bibitem[Delgado~Fuentealba et~al\mbox{.}(2021)]%
        {delgado_fuentealba_household_2021}
\bibfield{author}{\bibinfo{person}{Carlos~L. Delgado~Fuentealba},
  \bibinfo{person}{Jorge~A. Muñoz~Mendoza}, \bibinfo{person}{Sandra~M.
  Sepúlveda~Yelpo}, \bibinfo{person}{Carmen~L. Veloso~Ramos}, {and}
  \bibinfo{person}{Rodrigo~A. Fuentes-Solís}.}
  \bibinfo{year}{2021}\natexlab{}.
\newblock \showarticletitle{Household debt, automatic bill payments and
  inattention: {Theory} and evidence}.
\newblock \bibinfo{journal}{\emph{Journal of Economic Psychology}}
  \bibinfo{volume}{85} (\bibinfo{date}{Aug.} \bibinfo{year}{2021}),
  \bibinfo{pages}{102385}.
\newblock
\showISSN{01674870}
\urldef\tempurl%
\url{https://doi.org/10.1016/j.joep.2021.102385}
\showDOI{\tempurl}


\bibitem[Flitter(2020)]%
        {flitter_emily_their_2020}
\bibfield{author}{\bibinfo{person}{.~Emily Flitter}.}
  \bibinfo{year}{2020}\natexlab{}.
\newblock \showarticletitle{Their {Finances} {Ravaged}, {Customers} {Fear}
  {Banks} {Will} {Without} {Stimulus} {Checks}}.
\newblock \bibinfo{journal}{\emph{The New York Times}} (\bibinfo{date}{Dec.}
  \bibinfo{year}{2020}).
\newblock
\urldef\tempurl%
\url{https://www.nytimes.com/2020/12/31/business/stimulus-checks-overdraft.html}
\showURL{%
\tempurl}


\bibitem[Foster(2021)]%
        {foster_big_2021}
\bibfield{editor}{\bibinfo{person}{Ian Foster}} (Ed.).
  \bibinfo{year}{2021}\natexlab{}.
\newblock \bibinfo{booktitle}{\emph{Big data and social science: data science
  methods and tools for research and practice} (\bibinfo{edition}{second
  edition} ed.)}.
\newblock \bibinfo{publisher}{CRC Press}, \bibinfo{address}{Boca Raton, FL}.
\newblock
\showISBNx{978-0-367-56859-7 978-0-367-34187-9}


\bibitem[Friedman(2002)]%
        {friedman_stochastic_2002}
\bibfield{author}{\bibinfo{person}{Jerome~H. Friedman}.}
  \bibinfo{year}{2002}\natexlab{}.
\newblock \showarticletitle{Stochastic gradient boosting}.
\newblock \bibinfo{journal}{\emph{Computational Statistics \& Data Analysis}}
  \bibinfo{volume}{38}, \bibinfo{number}{4} (\bibinfo{date}{Feb.}
  \bibinfo{year}{2002}), \bibinfo{pages}{367--378}.
\newblock
\showISSN{01679473}
\urldef\tempurl%
\url{https://doi.org/10.1016/S0167-9473(01)00065-2}
\showDOI{\tempurl}


\bibitem[Jevtic et~al\mbox{.}(2022)]%
        {jevtic_ai_2022}
\bibfield{author}{\bibinfo{person}{Danijel Jevtic}, \bibinfo{person}{Romain
  Deleze}, {and} \bibinfo{person}{Joerg Osterrieder}.}
  \bibinfo{year}{2022}\natexlab{}.
\newblock \bibinfo{title}{{AI} for trading strategies}.
\newblock
\newblock
\urldef\tempurl%
\url{http://arxiv.org/abs/2208.07168}
\showURL{%
\tempurl}
\newblock
\shownote{arXiv:2208.07168 [q-fin]}.


\bibitem[Kumar et~al\mbox{.}(2022)]%
        {kumar_explainable_2022}
\bibfield{author}{\bibinfo{person}{Satyam Kumar}, \bibinfo{person}{Mendhikar
  Vishal}, {and} \bibinfo{person}{Vadlamani Ravi}.}
  \bibinfo{year}{2022}\natexlab{}.
\newblock \bibinfo{title}{Explainable {Reinforcement} {Learning} on {Financial}
  {Stock} {Trading} using {SHAP}}.
\newblock
\newblock
\urldef\tempurl%
\url{http://arxiv.org/abs/2208.08790}
\showURL{%
\tempurl}
\newblock
\shownote{arXiv:2208.08790 [cs]}.


\bibitem[Liu et~al\mbox{.}(2018)]%
        {liu_analyzing_2018}
\bibfield{author}{\bibinfo{person}{Xiao Liu}, \bibinfo{person}{Alan
  Montgomery}, {and} \bibinfo{person}{Kannan Srinivasan}.}
  \bibinfo{year}{2018}\natexlab{}.
\newblock \showarticletitle{Analyzing {Bank} {Overdraft} {Fees} with {Big}
  {Data}}.
\newblock \bibinfo{journal}{\emph{Marketing Science}} \bibinfo{volume}{37},
  \bibinfo{number}{6} (\bibinfo{date}{Nov.} \bibinfo{year}{2018}),
  \bibinfo{pages}{855--882}.
\newblock
\showISSN{0732-2399, 1526-548X}
\urldef\tempurl%
\url{https://doi.org/10.1287/mksc.2018.1106}
\showDOI{\tempurl}


\bibitem[Louppe et~al\mbox{.}(2013)]%
        {Louppe:2013:UVI:2999611.2999660}
\bibfield{author}{\bibinfo{person}{Gilles Louppe}, \bibinfo{person}{Louis
  Wehenkel}, \bibinfo{person}{Antonio Sutera}, {and} \bibinfo{person}{Pierre
  Geurts}.} \bibinfo{year}{2013}\natexlab{}.
\newblock \showarticletitle{Understanding Variable Importances in Forests of
  Randomized Trees}. In \bibinfo{booktitle}{\emph{Proceedings of the 26th
  International Conference on Neural Information Processing Systems}} (Lake
  Tahoe, Nevada) \emph{(\bibinfo{series}{NIPS'13})}. \bibinfo{publisher}{Curran
  Associates Inc.}, \bibinfo{address}{USA}, \bibinfo{pages}{431--439}.
\newblock
\urldef\tempurl%
\url{http://dl.acm.org/citation.cfm?id=2999611.2999660}
\showURL{%
\tempurl}


\bibitem[Melzer and Morgan(2015)]%
        {melzer_competition_2015}
\bibfield{author}{\bibinfo{person}{Brian~T. Melzer} {and}
  \bibinfo{person}{Donald~P. Morgan}.} \bibinfo{year}{2015}\natexlab{}.
\newblock \showarticletitle{Competition in a consumer loan market: {Payday}
  loans and overdraft credit}.
\newblock \bibinfo{journal}{\emph{Journal of Financial Intermediation}}
  \bibinfo{volume}{24}, \bibinfo{number}{1} (\bibinfo{date}{Jan.}
  \bibinfo{year}{2015}), \bibinfo{pages}{25--44}.
\newblock
\showISSN{10429573}
\urldef\tempurl%
\url{https://doi.org/10.1016/j.jfi.2014.07.001}
\showDOI{\tempurl}


\bibitem[Ramlal and Hosein(2021)]%
        {ramlal_personalized_2021}
\bibfield{author}{\bibinfo{person}{Karesia Ramlal} {and}
  \bibinfo{person}{Patrick Hosein}.} \bibinfo{year}{2021}\natexlab{}.
\newblock \showarticletitle{A {Personalized} {Overdraft} {Protection}
  {Framework}}. In \bibinfo{booktitle}{\emph{2021 {IEEE} 8th {International}
  {Conference} on {Data} {Science} and {Advanced} {Analytics} ({DSAA})}}.
  \bibinfo{publisher}{IEEE}, \bibinfo{address}{Porto, Portugal},
  \bibinfo{pages}{1--8}.
\newblock
\showISBNx{978-1-66542-099-0}
\urldef\tempurl%
\url{https://doi.org/10.1109/DSAA53316.2021.9564162}
\showDOI{\tempurl}


\bibitem[Rudin and Carlson(2019)]%
        {rudin_secrets_2019}
\bibfield{author}{\bibinfo{person}{Cynthia Rudin} {and} \bibinfo{person}{David
  Carlson}.} \bibinfo{year}{2019}\natexlab{}.
\newblock \showarticletitle{The {Secrets} of {Machine} {Learning}: {Ten}
  {Things} {You} {Wish} {You} {Had} {Known} {Earlier} to be {More} {Effective}
  at {Data} {Analysis}}.
\newblock  (\bibinfo{year}{2019}).
\newblock
\urldef\tempurl%
\url{https://doi.org/10.48550/ARXIV.1906.01998}
\showDOI{\tempurl}
\newblock
\shownote{Publisher: arXiv Version Number: 1}.


\bibitem[Saleiro et~al\mbox{.}(2018)]%
        {saleiro_aequitas_2018}
\bibfield{author}{\bibinfo{person}{Pedro Saleiro}, \bibinfo{person}{Benedict
  Kuester}, \bibinfo{person}{Loren Hinkson}, \bibinfo{person}{Jesse London},
  \bibinfo{person}{Abby Stevens}, \bibinfo{person}{Ari Anisfeld},
  \bibinfo{person}{Kit~T. Rodolfa}, {and} \bibinfo{person}{Rayid Ghani}.}
  \bibinfo{year}{2018}\natexlab{}.
\newblock \showarticletitle{Aequitas: {A} {Bias} and {Fairness} {Audit}
  {Toolkit}}.
\newblock  (\bibinfo{year}{2018}).
\newblock
\urldef\tempurl%
\url{https://doi.org/10.48550/ARXIV.1811.05577}
\showDOI{\tempurl}
\newblock
\shownote{Publisher: arXiv Version Number: 2}.


\bibitem[Stąpor(2018)]%
        {kurzynski_evaluating_2018}
\bibfield{author}{\bibinfo{person}{Katarzyna Stąpor}.}
  \bibinfo{year}{2018}\natexlab{}.
\newblock \showarticletitle{Evaluating and {Comparing} {Classifiers}: {Review},
  {Some} {Recommendations} and {Limitations}}.
\newblock In \bibinfo{booktitle}{\emph{Proceedings of the 10th {International}
  {Conference} on {Computer} {Recognition} {Systems} {CORES} 2017}},
  \bibfield{editor}{\bibinfo{person}{Marek Kurzynski}, \bibinfo{person}{Michal
  Wozniak}, {and} \bibinfo{person}{Robert Burduk}} (Eds.).
  Vol.~\bibinfo{volume}{578}. \bibinfo{publisher}{Springer International
  Publishing}, \bibinfo{address}{Cham}, \bibinfo{pages}{12--21}.
\newblock
\showISBNx{978-3-319-59161-2 978-3-319-59162-9}
\urldef\tempurl%
\url{https://doi.org/10.1007/978-3-319-59162-9_2}
\showDOI{\tempurl}
\newblock
\shownote{Series Title: Advances in Intelligent Systems and Computing}.


\bibitem[Tadapaneni(2019)]%
        {Tadapaneni2019-TADAII-2}
\bibfield{author}{\bibinfo{person}{Narendra~Rao Tadapaneni}.}
  \bibinfo{year}{2019}\natexlab{}.
\newblock \showarticletitle{Artificial Intelligence in Finance and
  Investments}.
\newblock \bibinfo{journal}{\emph{International Journal of Innovative Research
  in Science, Engineering and Technology}} \bibinfo{volume}{9},
  \bibinfo{number}{5} (\bibinfo{year}{2019}).
\newblock


\bibitem[Thaler and Sunstein(2021)]%
        {thaler_nudge_2021}
\bibfield{author}{\bibinfo{person}{Richard~H. Thaler} {and}
  \bibinfo{person}{Cass~R. Sunstein}.} \bibinfo{year}{2021}\natexlab{}.
\newblock \bibinfo{booktitle}{\emph{Nudge: the final edition}
  (\bibinfo{edition}{updated edition} ed.)}.
\newblock \bibinfo{publisher}{Penguin Books, an imprint of Penguin Random House
  LLC}, \bibinfo{address}{New York}.
\newblock
\showISBNx{978-0-14-313700-9}


\bibitem[Valenti(2022)]%
        {valenti_joe_overdraft_2022}
\bibfield{author}{\bibinfo{person}{Joe Valenti}.}
  \bibinfo{year}{2022}\natexlab{}.
\newblock \showarticletitle{Overdraft fees can price people out of banking}.
\newblock \bibinfo{journal}{\emph{Consumer Financial Protection Bureau}}
  (\bibinfo{date}{March} \bibinfo{year}{2022}).
\newblock
\urldef\tempurl%
\url{https://www.consumerfinance.gov/about-us/blog/overdraft-fees-can-price-people-out-of-banking/}
\showURL{%
\tempurl}


\bibitem[Varmedja et~al\mbox{.}(2019)]%
        {varmedja_credit_2019}
\bibfield{author}{\bibinfo{person}{Dejan Varmedja}, \bibinfo{person}{Mirjana
  Karanovic}, \bibinfo{person}{Srdjan Sladojevic}, \bibinfo{person}{Marko
  Arsenovic}, {and} \bibinfo{person}{Andras Anderla}.}
  \bibinfo{year}{2019}\natexlab{}.
\newblock \showarticletitle{Credit {Card} {Fraud} {Detection} - {Machine}
  {Learning} methods}. In \bibinfo{booktitle}{\emph{2019 18th {International}
  {Symposium} {INFOTEH}-{JAHORINA} ({INFOTEH})}}. \bibinfo{publisher}{IEEE},
  \bibinfo{address}{East Sarajevo, Bosnia and Herzegovina},
  \bibinfo{pages}{1--5}.
\newblock
\showISBNx{978-1-5386-7073-6}
\urldef\tempurl%
\url{https://doi.org/10.1109/INFOTEH.2019.8717766}
\showDOI{\tempurl}


\bibitem[Zhao et~al\mbox{.}(2022)]%
        {zhao_combining_2022}
\bibfield{author}{\bibinfo{person}{Yu Zhao}, \bibinfo{person}{Shaopeng Wei},
  \bibinfo{person}{Yu Guo}, \bibinfo{person}{Qing Yang},
  \bibinfo{person}{Xingyan Chen}, \bibinfo{person}{Qing Li},
  \bibinfo{person}{Fuzhen Zhuang}, \bibinfo{person}{Ji Liu}, {and}
  \bibinfo{person}{Gang Kou}.} \bibinfo{year}{2022}\natexlab{}.
\newblock \bibinfo{title}{Combining {Intra}-{Risk} and {Contagion} {Risk} for
  {Enterprise} {Bankruptcy} {Prediction} {Using} {Graph} {Neural} {Networks}}.
\newblock
\newblock
\urldef\tempurl%
\url{http://arxiv.org/abs/2202.03874}
\showURL{%
\tempurl}
\newblock
\shownote{arXiv:2202.03874 [cs, q-fin]}.


\end{thebibliography}

\appendix

\section{Appendix}

\begin{table}[H]
    \begin{tabular}{{p{3cm}p{5cm}}}
    \hline
        Model & Hyperparameters  \\ \hline
        Logistic Regression & penalty: [l1, l2"], C: [0.00001, 0.001, 0.1, 1, 10] \\ \hline
        Random Forest & n\_estimators: [100, 1000], max\_depth: [10, 50], max\_features: [sqrt, log2], min samples split: [2, 10] \\ \hline
        Gradient Boosted Decision Trees & nestimators: [100, 1000, 10000], learning rate: [0.01, 0.1, 0.5], subsample: [0.1, 0.5, 1.0], max\_depth: [5, 10] \\ \hline
        Decision Tree & criterion: [gini], max depth: [1, 5, 10, 20, 100] \\ \hline
        Feed Forward Neural Network & learning rate: [0.0001, 0.001, 0.01], hidden size: [128,256,512], dropout rate: [0.25], epochs: [10,30] \\ \hline
    \end{tabular}
    \caption{Model and Hyperparameter Grid for Model Search }
    \label{TABLE:modelgrid}
\end{table}

\begin{table}[H]
    \begin{tabular}{{p{2.2cm}p{.8cm}p{5cm}}}
    \hline
        Bank & Model  & Hyperparameters  \\ \hline
        Wells Fargo & GBDT & learning rate: 0.01, max depth: 10, estimators: 100, subsample: 0.5 \\ \hline
        Chase & GBDT & learning rate: 0.01, max depth: 10, estimators: 100, subsample: 0.5 \\ \hline
        Bank of America & GBDT & learning rate: 0.01, max depth: 10, estimators: 100, subsample: 0.5 \\ \hline
        Navy Federal Credit Union & GBDT & learning rate: 0.1, max depth: 5, estimators: 100, subsample: 0.5 \\ \hline
        US Bank & GBDT & learning rate: 0.01, max depth: 10, estimators: 100, subsample: 0.5 \\ \hline
        TD Bank & GBDT & learning rate: 0.01, max depth: 10, estimators: 100, subsample: 0.5 \\ \hline
        PNC Bank & GBDT & learning rate: 0.01, max depth: 10, estimators: 100, subsample: 0.5 \\ \hline
        Ally Financial  & FFNL & layers: 3, nodes: 128, dropout rate: 0.25, activation function: sigmoid, epochs 30, learning rate: 0.0001 \\ \hline
        Citibank & GBDT & learning rate: 0.01, max depth: 5, estimators: 100, subsample: 0.5 \\ \hline
    \end{tabular}
    \caption{Hyperparameters for production models per bank.}
    \label{TABLE:hyperparameters}
\end{table}

\begin{table}[H]
    \begin{tabular}{{p{4cm}p{4cm}}}
    \hline
        Bank & ROC AUC \\ \hline
        Wells Fargo & 0.84 \\ \hline
        Chase & 0.85 \\ \hline
        Bank of America & 0.85 \\ \hline
        Navy Federal Credit Union & 0.94 \\ \hline
        US Bank & 0.90 \\ \hline
        TD Bank & 0.86 \\ \hline
        PNC Bank & 0.92 \\ \hline
        Ally Financial & 0.81 \\ \hline
        Citibank & 0.86 \\ \hline
    \end{tabular}
    \caption{ROC AUC of production models per bank}
    \label{TABLE:ROCAUC}
\end{table}

\begin{table}[H]
\begin{tabular}{{p{4.5cm}p{1.5cm}p{1.5cm}}}
Classifier & Precision@k & Recall@k \\ \hline
Gradient Boosted Decision Trees & 0.45 & 0.49 \\ \hline
FFNL & 0.42 & 0.45 \\ \hline
Decision Tree & 0.36 & 0.38 \\ \hline
Random Forest & 0.44 & 0.47 \\ \hline
Logistic Regression & 0.42 & 0.44
\end{tabular}
\caption{Performance of Different Classifiers for Chase Bank}
\label{TABLE:ChasePerformance}
\end{table}

\begin{figure}[h]
	\includegraphics[width=\linewidth]{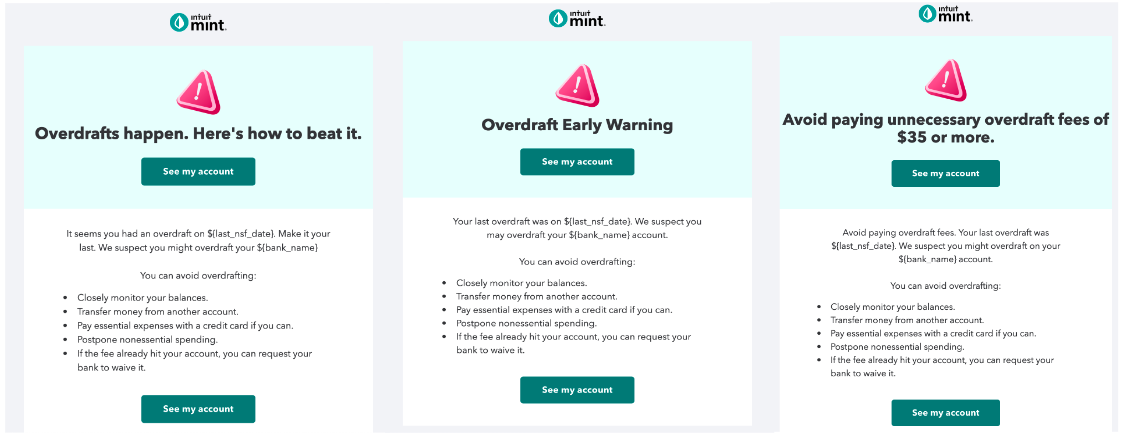}
    \caption{Overdraft Messages. Customers in the treatment group were sent a variety of messages meant to nudge them to change their behavior. This includes an empowering, baseline, and loss aversion message. The message also included the bank of the account we believed they were going to overdraft as well as the last date a customer had received an overdraft fee and a list of options for trying to avoid an overdraft fee. The baseline message, "Overdraft Early Warning", had the highest open-rate (30\%) compared to the other two messages. We speculate the simple message, "Overdraft Early Warning", better prompted the customer that they were at immediate risk of overdrawing their account compared to the other two messages. }
    \label{FIG:email}
\end{figure}

\end{document}